\pdfoutput=1

\documentclass[11pt]{article}

\usepackage[preprint]{acl}

\usepackage{times}
\usepackage{latexsym}

\usepackage[T1]{fontenc}

\usepackage[utf8]{inputenc}

\usepackage{microtype}

\usepackage{inconsolata}

\usepackage{graphicx}
\usepackage{booktabs}       
\usepackage{adjustbox}
\usepackage{amsmath}
\usepackage{mathtools}
\usepackage{algorithm}
\usepackage{algorithmic}
\usepackage{array}
\usepackage{enumitem}
\usepackage{multirow}
\usepackage{subfigure}
\setlist[enumerate]{nosep}
\usepackage{soul}
\usepackage{bm}
\usepackage{caption}
\usepackage{xcolor}
\usepackage{xspace}
\usepackage{arydshln}
\usepackage{bbding}

\definecolor{diff}{HTML}{D62F2F}
\newcommand{\diff}[1]{\textcolor{diff}{\small{(#1)}}}

\newcommand{\SAE}{en-US\xspace}

\newcommand{\AusE}{en-AU\xspace}

\newcommand{\IndE}{en-IN\xspace}

\newcommand{\BritE}{en-UK\xspace}

\newcommand{\bert}{\textsc{bert}\xspace}
\newcommand{\dis}{\textsc{distil}\xspace}
\newcommand{\rob}{\textsc{roberta}\xspace}

\newcommand{\simple}{\textsc{simple}\xspace}
\newcommand{\hard}{\textsc{hard}\xspace}

\title{Experiences from Creating a Benchmark for Sentiment Classification for Varieties of English}

\author{
 \textbf{Dipankar Srirag\textsuperscript{1}}\quad
 \textbf{Jordan Painter\textsuperscript{2}}\quad
 \textbf{Aditya Joshi\textsuperscript{1}}\quad
 \textbf{Diptesh Kanojia\textsuperscript{2}}
 \\
 \textsuperscript{1}University of New South Wales, Australia\\
 \textsuperscript{2}People-Centred AI, University of Surrey, United Kingdom\\
 \small{
   \textbf{Correspondence:} \href{mailto:d.srirag@unsw.edu.au}{\texttt{d.srirag@unsw.edu.au}}
 }
}

\begin{document}
\maketitle
\begin{abstract}
Existing benchmarks often fail to account for linguistic diversity, like language variants of English. In this paper, we share our experiences from our ongoing project of building a sentiment classification benchmark for three variants of English: Australian (en-AU), Indian (en-IN), and British (en-UK) English. Using Google Places reviews, we explore the effects of various sampling techniques based on label semantics, review length, and sentiment proportion and report performances on three fine-tuned BERT-based models. Our initial evaluation reveals significant performance variations influenced by sample characteristics, label semantics, and language variety, highlighting the need for nuanced benchmark design. We offer actionable insights for researchers to create robust benchmarks, emphasising the importance of diverse sampling, careful label definition, and comprehensive evaluation across linguistic varieties.
\end{abstract}
\section{Introduction}  
Benchmark-based evaluation of (large) language models is the prevalent norm in natural language processing (NLP) today~\cite{devlin-etal-2019-bert,achiam2023gpt,dubey2024llama}. However, these benchmarks do not adequately capture language variety evidenced through national/regional varieties, dialects, sociolects, or creoles\footnote{For the sake of brevity, we use the umbrella term `variety' when referring to three national varieties.}~\cite{jurgens2017incorporating,joshi2024naturallanguageprocessingdialects}. While several dialectal benchmarks exist for languages other than English~\cite{mdhaffar2017sentiment,al2018sentiment,oussous2020asa, faisal2024dialectbench}, limited work addresses the nuances of varieties of English. Existing approaches, such as Multi-VALUE~\cite{ziems2023multi}, primarily cater to using American English test sets to evaluate for language varieties and do not fully capture the complex interplay of vocabulary, syntax, and cultural knowledge unique to different English-speaking communities. Despite the superlative performance reported by language models on Standard American English datasets, there is a significant gap in evaluating how well these models perform across different varieties of English, such as Australian (en-AU), Indian (en-IN), and British (en-UK) English. Recognising this gap, this work is part of a broader effort to develop benchmarks that reflect these varieties more comprehensively. Rather than presenting a completed benchmark, we share our experiences and insights gained from designing a sentiment classification benchmark tailored to these linguistic variations. Specifically, we address the question:
\begin{quote}
    \textit{``What data sampling strategies can be employed to create a benchmark for sentiment classification for different varieties of English?''}
\end{quote}

To explore this, we collected self-supervised Google Places reviews and employed various sampling strategies. Our initial findings highlight the significant impact of these sampling techniques on model performance, revealing nuanced variations influenced by sample characteristics and language variety. By evaluating three BERT-based models fine-tuned on this data, we uncover key insights that can inform the design of future benchmarks.

With the evaluation of language varieties, this paper joins the recent NLP research that calls for robustness towards dialects and language variations in NLP modeling and evaluation~\cite{demszky-etal-2021-learning, wang-etal-2022-measure, blaschke-etal-2024-dialect, holt-etal-2024-perceptions}. We offer actionable recommendations for researchers and practitioners, advocating for diverse sampling approaches, and thoughtful label definitions, supported by comprehensive performance evaluation to ensure that NLP models are tested on generalisable and robust benchmarks.
\section{Methodology}
\begin{figure}[t!]
    \centering
    \includegraphics[width=1\linewidth]{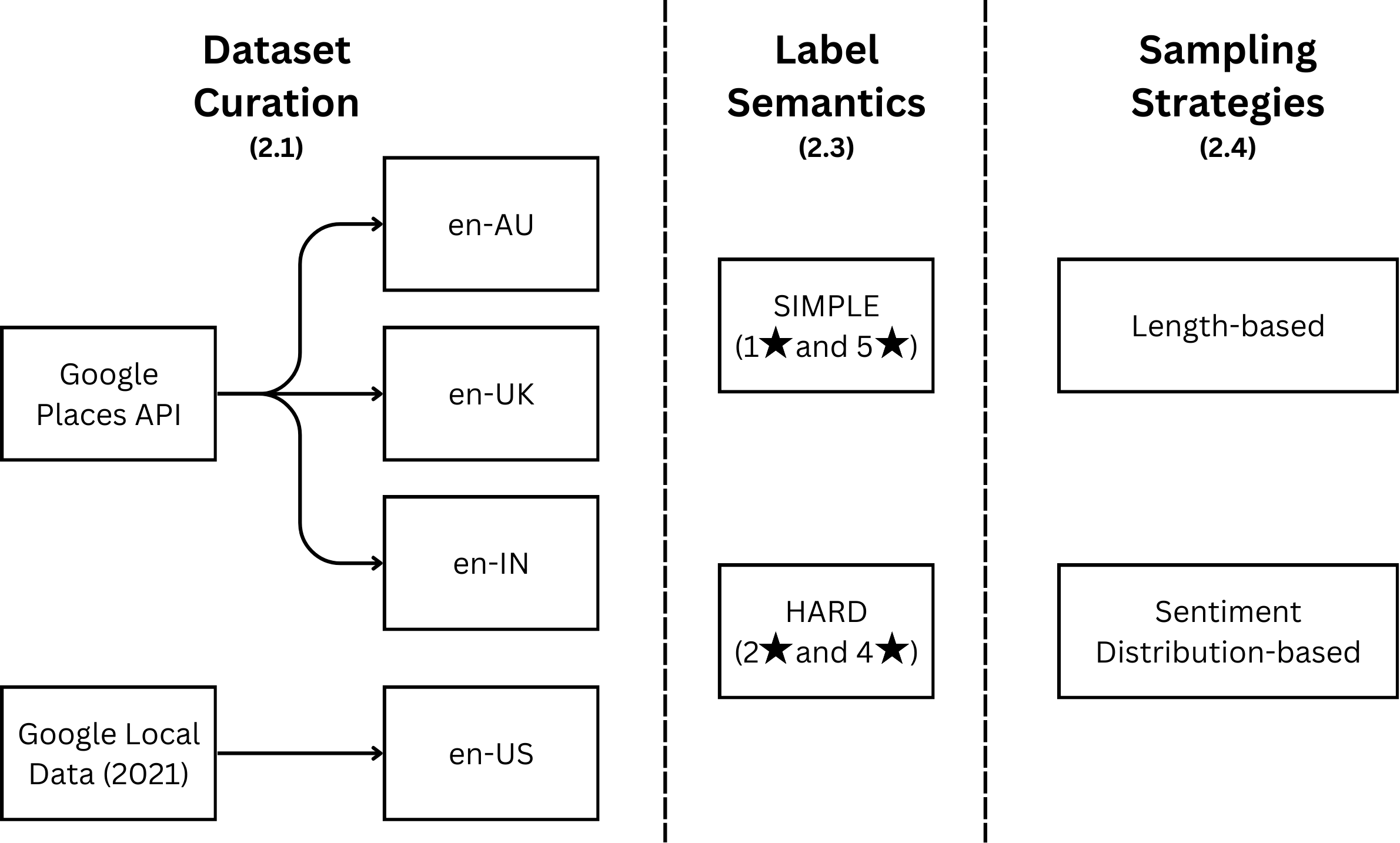}
    \caption{Data Curation and Sampling for Sentiment Classification for varieties of English.}
    \label{fig:method}
\end{figure}

Our study focused on boolean sentiment classification of Google Places Reviews, leveraging self-supervised sentiment labels derived from user ratings. Figure~\ref{fig:method} illustrates the methodology we used for data curation and sampling.

\subsection{Dataset}
We collect reviews and their corresponding star ratings from the Google Places API\footnote{\textit{\url{https://developers.google.com/maps/documentation/places/web-service/overview}}; Accessed on 23 July 2024} from the place types, also defined by the API. We apply location-based filtering to curate datasets from three locales: Australia (en-AU), India (en-IN), and the United Kingdom (en-UK).

We base our city selection criteria on country-specific population thresholds: cities with at least 10,000 residents in Australia\footnote{Australian Bureau of Statistics. \textit{\url{https://www.abs.gov.au/statistics/standards}}; Accessed on 29 July 2024}, 100,000 residents in India\footnote{Census India. \textit{\url{http://www.censusindia.gov.in/Metadata/Metada.htm\#2c}}; Accessed on 29 July 2024}. In the United Kingdom, due to the complex nature of defining a city, we select geographic locations with populations exceeding $50,000$ residents. Additionally, we incorporate a subset of 29,512 reviews from New York City, as used in previous work~\cite{li-etal-2022-uctopic,10.1145/3539618.3592036}, to establish a baseline for Standard American English (en-US). 

We additionally filter non-English text, thresholding language probabilities derived from fastText~\cite{grave2018learning} word vectors. Table~\ref{tab:processed_rats} provides an overview of the review counts and average language probabilities for each locale. We do not collect or store information about the reviewer or the place to ensure the anonymity of the reviews. We pre-process the reviews to remove special characters and emoticons.

\begin{table}[]
    \begin{adjustbox}{width=1\linewidth,center}
        \begin{tabular}{ccccccc}
            \toprule
            Locale & 1 \FiveStar & 2 \FiveStar & 3 \FiveStar & 4 \FiveStar & 5 \FiveStar & $\bm{\mu}$\\\midrule[\heavyrulewidth]
             \SAE & 5351 & 2389 & 1612 & 5142 & 15097 & 0.985\\
             \AusE & 924 & 359 & 510 & 1108 & 4570 & 0.998\\
             \BritE & 2691 & 955 & 1495 & 3399 & 13430 & 0.999\\
             \IndE & 4268 & 1306 & 3309 & 7127 & 15629 & 0.988\\
             \bottomrule
        \end{tabular}
    \end{adjustbox}
    \caption{Review counts across different ratings for Standard American English (\SAE), Australian English (\AusE), British English (\BritE), and Indian English (\IndE) locales. $\bm{\mu}$ is the average language probability (calculated using fastText~\cite{grave2018learning} word vectors) of a review being in English.}
    \label{tab:processed_rats}
\end{table}

\subsection{Data Quality and Validation}
The assumption that location-based filtering would yield datasets representative of the local linguistic varieties is supported by prior research evaluating NLP robustness to dialectal variation~\cite{blodgett-etal-2016-demographic,sun-etal-2023-dialect}.  However, we recognize that there is limited research specifically validating the representativeness of user-generated location-based reviews. Additionally, to mitigate the risk of including reviews authored by tourists rather than local residents, we excluded reviews from locations designated as \textit{`tourist attractions'}\footnote{\textit{\url{https://developers.google.com/maps/documentation/places/web-service/supported_types}}; Accessed on 13 November 2024.} by the Google Places API.

To further validate the representativeness of our data, two authors of this paper—one a native speaker of British English and the other of Indian English, manually labeled a sample of reviews as en-AU, en-IN, en-UK, or `Cannot say' for ambiguous cases. The agreement rates with the original locale labels were 0.64 and 0.88 for en-IN reviews, while Cohen’s kappa between the annotators was only 0.191, indicating low inter-annotator agreement, particularly for inner-circle English varieties. A detailed analysis (provided in Appendix~\ref{sec:annot}) showed greater annotation consistency for en-IN but significant overlap between en-AU and en-UK, highlighting the complexity of locale identification.

\subsection{Label Semantics}
Google Places Reviews contain user-defined ratings of 1 to 5, where 5 is the highest. For the task of Boolean sentiment classification, we selected a strategy to define what a `positive' or `negative' review means, \textit{i.e.}, label semantics. We compared two label semantic configurations: subsets with well-separated labels ($1$\FiveStar and $5$\FiveStar; \simple) and subsets with closer labels ($2$\FiveStar and $4$\FiveStar; \hard). We acknowledge that these are not the only ways to convert a dataset with $5$ ratings into a boolean classification task. However, the goal here is to measure performance based on the separation of labels (hence, the names \simple and \hard) to build a challenging benchmark. Dataset statistics are in Table~\ref{tab:processed_rats} with detailed tables in the Appendix.

\subsection{Sampling Strategies}\label{sec:strategies}
\paragraph{Length-based sampling}
We examine the influence of review length by splitting reviews in terms of the number of words. Given a set of all words in a review ($\mathcal{R}$), each subset is sampled as follows: (A) \textsc{len-75}: Reviews with lengths greater than the first quartile, indicating top 75\% longest reviews; (B) \textsc{len-50}: Reviews with lengths greater than the second quartile, indicating top 50\% longest reviews; (C) \textsc{len-25}: Reviews with lengths greater than the third quartile, indicating top 25\% longest reviews.

\paragraph{Sentiment Distribution-based sampling}
To identify sentiment-bearing words in each review, we used the spaCy\footnote{\textit{\url{https://spacy.io/universe/project/spacy-textblob}}; Accessed on 29 July 2024.} to process each review and count the number of positive and negative words based on their polarity scores. Words with a positive polarity were counted as positive sentiment-bearing words (\textsc{pos}), while those with a negative polarity were counted as negative sentiment-bearing words (\textsc{neg}). The sentiment distribution $\rho$ was then computed as the proportion of sentiment-bearing words in a review, and is computed as follows:
\begin{align}
    \rho = \frac{|\textsc{pos}| + |\textsc{neg}|}{|\mathcal{R}|}
    \nonumber
\end{align}
Here, $|\cdot|$ is the number of words in a set. We then create each subset as follows: (A) \textsc{sent-75}: Reviews with sentiment distribution greater than the first quartile, indicating top $75$\% reviews in terms of proportion of sentiment-bearing words. (B) \textsc{sent-50}: Reviews with sentiment distribution greater than the second quartile, indicating top $50$\% reviews in terms of proportion of sentiment-bearing words; (C) \textsc{sent-25}: Reviews with sentiment distribution greater than the third quartile, indicating top $25$\% reviews in terms of proportion of sentiment-bearing words.


\section{Experiment Setup}
\begin{table}[h!]
    \begin{adjustbox}{width=0.4\textwidth,center}
        \setlength{\cmidrulekern}{0.25em}
        \begin{tabular}{cccccc}
            \toprule
            \multicolumn{6}{c}{Length-based}\\\midrule[\heavyrulewidth]
            \multirow{2}{4em}{\centering Locale} & \multirow{2}{4em}{\centering Sample} & \multicolumn{4}{c}{Rating}\\\cmidrule(lr){3-6}
            & & 1 \FiveStar & 5 \FiveStar & 2 \FiveStar & 4 \FiveStar\\\midrule[\heavyrulewidth]
             \multirow{4}{4em}{\SAE}%
             & \textsc{len-75} & 4390 & 10137 & 1949 & 3606\\
             & \textsc{len-50} & 3638 & 6401 & 1544 & 2295\\
             & \textsc{len-25} & 2373 & 2547 & 884 & 953\\\hdashline
             \multirow{4}{4em}{\AusE}%
             & \textsc{len-75} & 826 & 3218 & 305 & 828\\
             & \textsc{len-50} & 641 & 1995 & 233 & 575\\
             & \textsc{len-25} & 372 & 889 & 142 & 307\\\hdashline
             \multirow{4}{4em}{\BritE}%
             & \textsc{len-75} & 2376 & 9560 & 842 & 2653\\
             & \textsc{len-50} & 1834 & 5918 & 659 & 1807\\
             & \textsc{len-25} & 1120 & 2615 & 401 & 891\\\hdashline
             \multirow{4}{4em}{\IndE}%
             & \textsc{len-75} & 3675 & 10903 & 1081 & 5676\\
             & \textsc{len-50} & 2728 & 6682 & 811 & 3841\\
             & \textsc{len-25} & 1429 & 3123 & 452 & 1977\\\midrule[\heavyrulewidth]
             \multicolumn{6}{c}{Sentiment Distribution-based}\\\midrule[\heavyrulewidth]
             \multirow{2}{4em}{\centering Locale} & \multirow{2}{4em}{\centering Sample} & \multicolumn{4}{c}{Rating}\\\cmidrule(lr){3-6}
            & & 1 \FiveStar & 5 \FiveStar & 2 \FiveStar & 4 \FiveStar\\\midrule[\heavyrulewidth]
             \multirow{4}{4em}{\SAE}%
             & \textsc{sent-75} & 2538 & 11977 & 1378 & 4127\\
             & \textsc{sent-50} & 1336 & 8818 & 755 & 2992\\
             & \textsc{sent-25} & 632 & 4819 & 318 & 1638\\\hdashline
             \multirow{4}{4em}{\AusE}%
             & \textsc{sent-75} & 386 & 3787 & 202 & 888\\
             & \textsc{sent-50} & 169 & 2660 & 110 & 597\\
             & \textsc{sent-25} & 66 & 1396 & 42 & 295\\\hdashline
             \multirow{4}{4em}{\BritE}%
             & \textsc{sent-75} & 1055 & 11242 & 528 & 2686\\
             & \textsc{sent-50} & 460 & 7969 & 265 & 1752\\
             & \textsc{sent-25} & 159 & 4221 & 100 & 823\\\hdashline
             \multirow{4}{4em}{\IndE}%
             & \textsc{sent-75} & 2659 & 12665 & 846 & 5564\\
             & \textsc{sent-50} & 1455 & 9000 & 246 & 1692\\
             & \textsc{sent-25} & 686 & 5452 & 447 & 3503\\
             \bottomrule
        \end{tabular}
    \end{adjustbox}
    \caption{Counts of reviews by rating (1\FiveStar, 2\FiveStar, 4\FiveStar, 5\FiveStar) in length-based and sentiment distribution-based samples across different locales.}
    \label{tab:processed_stats}
\end{table}
We conduct experiments using three encoder models: BERT-base~\cite{devlin-etal-2019-bert}, DistilBERT-base~\cite{sanh2020distilbertdistilledversionbert}, and RoBERTa-base~\cite{zhuang-etal-2021-robustly}. For each language variety and sampling strategy described in Section~\ref{sec:strategies}, we divide the data into training, validation, and test sets in proportions of 70\%, 15\%, and 15\%, respectively. Table~\ref{tab:processed_stats} summarises the statistics of each sample across the different language varieties and sampling strategies.

We perform `\textit{in-sample}' fine-tuning, \textit{i.e.} a model is trained and evaluated on subsets derived from the same sample. We train the models for a maximum of 10 epochs, employing early stopping to avoid overfitting. The training is optimized using the Adam optimiser, and all experiments are run on a setup with six NVIDIA V100 GPUs. To ensure reproducibility, we use a fixed random seed for all experiments.

We evaluate model performance using three macro-averaged metrics: Precision (P), Recall (R), and F1-score (F), all expressed as percentages.


\section{Evaluation}
Our evaluation examines the effects of label semantics, review length, and sentiment distribution on model performance across different English language varieties. The main findings are summarized below, reporting both model-specific and aggregate results.

\subsection{Impact of Label Semantics}
Table~\ref{tab:rating_detailed} compares model performance on samples from each locale when using the \simple and \hard label semantic configurations. Models fine-tuned on \simple samples of en-AU achieved the highest average F1-score of 97.2, while the lowest average F1-score of 93.5 was reported for models trained on en-US. However, switching from \simple to \hard labels resulted in a significant performance drop across all locales, with an average F1-score decrease of 13.7 points. The highest degradation is seen in models trained on en-IN samples (19.4 points), with models trained on en-US samples being the least affected (7.0 points). This outcome confirms the sensitivity of model performance to label semantics, reinforcing the need for careful label definition when designing benchmarks.
\begin{table*}[t!]
    \begin{adjustbox}{width=1\linewidth,center}
        \renewcommand{\arraystretch}{1.1}
        \setlength{\tabcolsep}{7pt}
        \setlength{\cmidrulekern}{0.3em}
        \setlength{\cmidrulekern}{0.3em}
        \begin{tabular}{m{4em}m{4em}cccccccccccc}
        \toprule
        \multirow{2}{4em}{Locale} & \multirow{2}{4em}{Label} & \multicolumn{3}{c}{\bert} & \multicolumn{3}{c}{\dis} & \multicolumn{3}{c}{\rob} & \multicolumn{3}{c}{$\mu$}\\\cmidrule(lr){3-5}\cmidrule(lr){6-8}\cmidrule(lr){9-11}\cmidrule(lr){12-14}
        & & P & R & F & P & R & F & P & R & F & P & R & F\\\midrule[\heavyrulewidth]
        \multirow{2}{4em}{\SAE} %
        & \simple & 93.3 & 94.9 & 94.1 & 92.6 & 95.2 & 93.8 & 92.2 & 92.8 & 92.5 & 92.7 & 94.3 & 93.5\\
        & \hard & 85.4 & 87.4 & 86.3 & 83.0 & 86.0 & 84.0 & 88.1 & 90.6 & 89.1 & 85.5 & 88.0 & 86.5 \diff{7.0}\\\hdashline
        \multirow{2}{4em}{\AusE} %
        & \simple & 92.7 & 98.2 & 95.2 & 97.0 & 97.8 & 97.4 & 98.8 & 99.2 & 99.0 & 96.2 & 98.4 & 97.2\\
        & \hard & 81.0 & 73.7 & 76.2 & 78.1 & 85.0 & 79.9 & 88.1 & 90.8 & 89.4 & 82.4 & 83.1 & 81.8 \diff{15.4}\\\hdashline
        \multirow{2}{4em}{\BritE} %
        & \simple & 96.0 & 97.9 & 97.0 & 96.9 & 95.5 & 96.2 & 97.1 & 97.8 & 97.5 & 96.7 & 97.1 & 96.9\\
        & \hard & 82.8 & 88.7 & 85.1 & 79.5 & 88.7 & 81.9 & 82.8 & 90.6 & 85.5 & 81.7 & 89.3 & 84.2 \diff{12.7}\\\hdashline
        \multirow{2}{4em}{\IndE} %
        & \simple & 97.3 & 95.5 & 96.4 & 95.4 & 94.3 & 94.9 & 92.2 & 95.6 & 93.8 & 95.0 & 95.1 & 95.0\\
        & \hard & 77.1 & 73.5 & 75.1 & 75.4 & 77.4 & 76.3 & 79.9 & 72.4 & 75.3 & 77.5 & 74.4 & 75.6 \diff{19.4}\\
        \bottomrule
        \end{tabular}
    \end{adjustbox}
    \caption{Performance comparison between different label semantics. $\mu$ is the average performance computed across all models on samples from a locale. The average performance degradation reported by models on samples with \hard labels compared to \simple labels is represented by the \textcolor{diff}{(number)}.}
    \label{tab:rating_detailed}
\end{table*}

\subsection{Impact of Sampling Strategies}
\paragraph{Length-based sampling} Table~\ref{tab:len_detailed} illustrates the trend in F1-scores, averaged across all models, when fine-tuned on subsets sampled based on review lengths from each locale. Models trained on \textsc{simple-len-75} samples of en-AU performed the best, with an average F1-score of 98.3. In contrast, the worst performance was observed for models trained on \textsc{hard-len-25} samples of en-IN, with an average F1-score of 81.5. The results suggest that models benefit from longer reviews in most cases, especially when trained on samples from inner-circle varieties like en-AU and en-UK. However, the performance of models trained on Indian English reviews was consistently lower across different review lengths, indicating that linguistic variations may play a role in model robustness.

\begin{table*}[t!]
    \begin{adjustbox}{width=1\linewidth,center}
        \renewcommand{\arraystretch}{1.1}
        \setlength{\tabcolsep}{9pt}
        \setlength{\cmidrulekern}{0.3em}
        \begin{tabular}{m{4em}m{4em}cccccccccccc}
        \toprule
        \multicolumn{14}{c}{\simple}\\\midrule[\heavyrulewidth]
        \multirow{2}{4em}{Locale} & \multirow{2}{4em}{Sample} & \multicolumn{3}{c}{\bert} & \multicolumn{3}{c}{\dis} & \multicolumn{3}{c}{\rob} & \multicolumn{3}{c}{$\mu$}\\\cmidrule(lr){3-5}\cmidrule(lr){6-8}\cmidrule(lr){9-11}\cmidrule(lr){12-14}
        & & P & R & F & P & R & F & P & R & F & P & R & F\\\midrule[\heavyrulewidth]
        \multirow{3}{4em}{\SAE} %
            & \textsc{len-75} & 92.7 & 94.1 & 93.3 & 94.5 & 94.8 & 94.6 & 91.9 & 95.1 & 93.3 & 93.0 & 94.7 & 93.7 \\
            & \textsc{len-50} & 97.1 & 96.4 & 96.7 & 94.5 & 95.6 & 95.0 & 96.1 & 96.9 & 96.5 & 95.9 & 96.3 & 96.1 \\
            & \textsc{len-25} & 95.3 & 95.3 & 95.3 & 96.4 & 96.3 & 96.3 & 97.0 & 97.0 & 97.0 & 96.2 & 96.2 & 96.2 \\\hdashline
        \multirow{3}{4em}{\AusE} %
            & \textsc{len-75} & 96.6 & 99.1 & 97.8 & 97.7 & 99.4 & 98.5 & 97.7 & 99.4 & 98.5 & 97.3 & 99.3 & 98.3 \\
            & \textsc{len-50} & 89.1 & 95.0 & 91.4 & 97.4 & 96.4 & 96.9 & 96.1 & 98.0 & 97.0 & 94.2 & 96.4 & 95.1 \\
            & \textsc{len-25} & 98.1 & 98.1 & 98.1 & 96.2 & 96.2 & 96.2 & 98.7 & 99.4 & 99.1 & 97.7 & 97.9 & 97.8 \\\hdashline
        \multirow{3}{4em}{\BritE} %
            & \textsc{len-75} & 97.7 & 97.8 & 97.8 & 95.9 & 97.7 & 96.8 & 95.7 & 98.3 & 96.9 & 96.4 & 98.0 & 97.2 \\
            & \textsc{len-50} & 96.6 & 98.2 & 97.4 & 97.9 & 98.6 & 98.2 & 97.8 & 99.1 & 98.4 & 97.4 & 98.6 & 98.0 \\
            & \textsc{len-25} & 95.3 & 97.5 & 96.3 & 95.4 & 96.5 & 95.9 & 97.5 & 97.5 & 97.5 & 96.0 & 97.1 & 96.5 \\\hdashline
        \multirow{3}{4em}{\IndE} %
            & \textsc{len-75} & 97.0 & 96.2 & 96.6 & 95.1 & 95.3 & 95.2 & 99.2 & 98.8 & 99.0 & 97.1 & 96.8 & 96.9 \\
            & \textsc{len-50} & 98.1 & 98.0 & 98.1 & 96.5 & 97.0 & 96.8 & 97.0 & 98.0 & 97.4 & 97.2 & 97.7 & 97.4 \\
            & \textsc{len-25} & 98.8 & 99.2 & 99.0 & 97.6 & 97.8 & 97.7 & 96.5 & 95.5 & 96.0 & 97.6 & 97.5 & 97.6 \\
        \midrule[\heavyrulewidth]
        \multicolumn{14}{c}{\hard}\\\midrule[\heavyrulewidth]
        \multirow{2}{4em}{Locale} & \multirow{2}{4em}{Sample} & \multicolumn{3}{c}{\bert} & \multicolumn{3}{c}{\dis} & \multicolumn{3}{c}{\rob} & \multicolumn{3}{c}{$\mu$}\\\cmidrule(lr){3-5}\cmidrule(lr){6-8}\cmidrule(lr){9-11}\cmidrule(lr){12-14}
        & & P & R & F & P & R & F & P & R & F & P & R & F\\\midrule[\heavyrulewidth]
        \multirow{3}{4em}{\SAE} %
            & \textsc{len-75} & 87.3 & 88.3 & 87.7 & 82.3 & 85.1 & 82.8 & 89.5 & 87.7 & 88.5 & 86.4 & 87.1 & 86.3 \\
            & \textsc{len-50} & 85.8 & 84.8 & 85.2 & 86.4 & 85.9 & 86.1 & 90.2 & 90.4 & 90.3 & 87.5 & 87.0 & 87.2 \\
            & \textsc{len-25} & 87.9 &85.8 & 85.2 & 87.5 & 86.7 & 86.4 & 91.4 & 91.0 & 90.7 & 88.9 & 87.8 & 87.4 \\\hdashline
        \multirow{3}{4em}{\AusE} %
            & \textsc{len-75} & 88.1 & 87.3 & 87.7 & 88.0 & 84.7 & 86.1 & 87.2 & 89.3 & 88.2 & 87.8 & 87.1 & 87.3 \\
            & \textsc{len-50} & 84.5 & 78.6 & 80.7 & 80.9 & 80.2 & 80.5 & 85.8 & 84.0 & 84.8 & 83.8 & 80.9 & 82.0 \\
            & \textsc{len-25} & 86.4 & 90.0 & 87.7 & 86.4 & 90.0 & 87.7 & 88.9 & 91.6 & 90.0 & 87.3 & 90.5 & 88.5 \\\hdashline
        \multirow{3}{4em}{\BritE} %
            & \textsc{len-75} & 84.2 & 89.2 & 86.2 & 83.8 & 86.4 & 84.9 & 82.3 & 88.9 & 84.5 & 83.4 & 88.1 & 85.2 \\
            & \textsc{len-50} & 86.8 & 87.8 & 87.3 & 84.7 & 77.3 & 79.9 & 83.9 & 84.2 & 84.1 & 85.1 & 83.1 & 83.7 \\
            & \textsc{len-25} & 94.0 & 93.3 & 93.6 & 90.1 & 93.1 & 91.3 & 96.4 & 96.4 & 96.4 & 93.5 & 94.3 & 93.8 \\\hdashline
        \multirow{3}{4em}{\IndE} %
            & \textsc{len-75} & 80.0 & 80.9 & 80.5 & 80.1 & 82.5 & 81.2 & 81.4 & 89.1 & 84.3 & 80.5 & 84.2 & 82.0 \\
            & \textsc{len-50} & 85.2 & 81.3 & 83.0 & 76.9 & 83.2 & 79.3 & 87.8 & 86.3 & 87.0 & 83.3 & 83.6 & 83.1 \\
            & \textsc{len-25} & 84.4 & 86.7 & 85.5 & 82.1 & 77.5 & 79.5 & 83.7 & 76.5 & 79.4 & 83.4 & 80.2 & 81.4 \\
        \bottomrule
        \end{tabular}
    \end{adjustbox}
    \caption{Performance comparison of models trained on samples created based on review length. $\mu$ is the average performance computed across all models.}
    \label{tab:len_detailed}
\end{table*}

\paragraph{Sentiment Distribution-based Sampling}
Table~\ref{tab:pol_detailed} shows the F1-scores, averaged across all models, for sentiment distribution-based sampling. Models trained on \textsc{simple-sent-75} samples of en-AU once again outperformed others, with an average F1-score of 97.5. As sentiment proportion increased in the \hard samples, performance degraded significantly, with the lowest F1-score of 64.8 reported for models trained on \textsc{hard-sent-25} samples. This degradation suggests that higher densities of sentiment-bearing words introduce ambiguity, making sentiment classification more challenging for the models. While models trained on inner-circle varieties exhibited performance drops as sentiment density increased, models trained on en-IN reviews showed more pronounced variations, reflecting the linguistic diversity inherent in this variety.

\begin{table*}[t!]
    \begin{adjustbox}{width=1\linewidth,center}
        \renewcommand{\arraystretch}{1.1}
        \setlength{\tabcolsep}{9pt}
        \setlength{\cmidrulekern}{0.3em}
        \begin{tabular}{m{4em}m{4em}cccccccccccc}
        \toprule
        \multicolumn{14}{c}{\simple}\\\midrule[\heavyrulewidth]
        \multirow{2}{4em}{Locale} & \multirow{2}{4em}{Sample} & \multicolumn{3}{c}{\bert} & \multicolumn{3}{c}{\dis} & \multicolumn{3}{c}{\rob} & \multicolumn{3}{c}{$\mu$}\\\cmidrule(lr){3-5}\cmidrule(lr){6-8}\cmidrule(lr){9-11}\cmidrule(lr){12-14}
        & & P & R & F & P & R & F & P & R & F & P & R & F\\\midrule[\heavyrulewidth]
        \multirow{3}{4em}{\SAE} %
        & \textsc{sent-75} & 92.1 & 94.2 & 93.1 & 92.3 & 95.2 & 93.6 & 95.3 & 96.1 & 95.7 & 93.2 & 95.2 & 94.1\\
        & \textsc{sent-50} & 90.3 & 95.1 & 92.5 & 91.8 & 96.4 & 93.9 & 94.0 & 96.5 & 95.1 & 92.0 & 96.0 & 93.8\\
        & \textsc{sent-25} & 95.6 & 92.4 & 94.0 & 94.4 & 93.8 & 94.1 & 95.2 & 93.9 & 94.5 & 95.1 & 93.4 & 94.2\\\hdashline
        \multirow{3}{4em}{\AusE} %
        & \textsc{sent-75} & 95.3 & 99.5 & 97.3 & 94.3 & 99.3 & 96.7 & 98.6 & 98.6 & 98.6 & 96.1 & 99.1 & 97.5\\
        & \textsc{sent-50} & 99.8 & 97.1 & 98.3 & 91.9 & 96.5 & 94.1 & 100.0 & 100.0 & 100.0 & 97.2 & 97.9 & 97.5\\
        & \textsc{sent-25} & 88.9 & 99.2 & 93.4 & 100.0 & 100.0 & 100.0 & 93.8 & 99.6 & 96.5 & 94.2 & 99.6 & 96.6\\\hdashline
        \multirow{3}{4em}{\BritE} %
        & \textsc{sent-75} & 97.5 & 94.6 & 96.0 & 95.8 & 93.1 & 94.4 & 99.2 & 91.0 & 94.7 & 97.5 & 92.9 & 95.0\\
        & \textsc{sent-50} & 94.9 & 98.6 & 96.7 & 90.1 & 98.2 & 93.7 & 94.9 & 98.6 & 96.7 & 93.3 & 98.5 & 95.7\\
        & \textsc{sent-25} & 99.8 & 93.8 & 96.5 & 96.8 & 96.8 & 96.8 & 90.0 & 99.5 & 94.2 & 95.5 & 96.7 & 95.8\\\hdashline
        \multirow{3}{4em}{\IndE} %
        & \textsc{sent-75} & 96.6 & 94.6 & 95.6 & 95.1 & 95.9 & 95.5 & 86.8 & 94.8 & 90.4 & 92.8 & 95.1 & 93.8\\
        & \textsc{sent-50} & 94.0 & 92.7 & 93.4 & 95.4 & 93.3 & 94.3 & 87.3 & 92.8 & 89.8 & 92.2 & 92.9 & 92.5\\
        & \textsc{sent-25} & 93.8 & 89.3 & 91.4 & 96.3 & 90.3 & 93.0 & 90.3 & 93.9 & 92.0 & 93.5 & 91.2 & 92.1\\\midrule[\heavyrulewidth]
        \multicolumn{14}{c}{\hard}\\\midrule[\heavyrulewidth]
        \multirow{2}{4em}{Locale} & \multirow{2}{4em}{Sample} & \multicolumn{3}{c}{\bert} & \multicolumn{3}{c}{\dis} & \multicolumn{3}{c}{\rob} & \multicolumn{3}{c}{$\mu$}\\\cmidrule(lr){3-5}\cmidrule(lr){6-8}\cmidrule(lr){9-11}\cmidrule(lr){12-14}
        & & P & R & F & P & R & F & P & R & F & P & R & F\\\midrule[\heavyrulewidth]
        \multirow{3}{4em}{\SAE} %
        & \textsc{sent-75} & 86.8 & 85.1 & 85.9 & 85.0 & 85.4 & 85.9 & 85.5 & 90.6 & 87.4 & 85.8 & 87.0 & 86.4\\
        & \textsc{sent-50} & 86.1 & 82.0 & 83.8 & 86.8 & 86.8 & 86.8 & 87.1 & 87.5 & 87.3 & 86.7 & 85.4 & 86.0\\
        & \textsc{sent-25} & 88.2 & 87.3 & 87.8 & 86.4 & 85.4 & 85.4 & 90.0 & 92.0 & 90.9 & 88.2 & 88.2 & 88.0\\\hdashline
        \multirow{3}{4em}{\AusE} %
        & \textsc{sent-75} & 81.9 & 75.2 & 77.9 & 75.6 & 79.4 & 63.4 & 83.5 & 82.2 & 82.8 & 80.3 & 74.7 & 78.9\\
        & \textsc{sent-50} & 85.6 & 80.2 & 82.5 & 87.1 & 71.9 & 76.6 & 83.5 & 75.6 & 78.7 & 85.4 & 75.9 & 79.3\\
        & \textsc{sent-25} & 95.5 & 62.5 & 67.6 & 70.3 & 60.8 & 63.4 & 70.3 & 60.8 & 63.4 & 78.7 & 61.4 & 64.8\\\hdashline
        \multirow{3}{4em}{\BritE} %
        & \textsc{sent-75} & 88.7 & 88.7 & 88.7 & 83.8 & 88.2 & 85.7 & 89.1 & 89.7 & 89.4 & 87.2 & 88.9 & 87.9\\
        & \textsc{sent-50} & 88.0 & 85.6 & 86.8 & 84.0 & 88.2 & 85.9 & 87.2 & 90.6 & 88.8 & 86.4 & 88.1 & 87.2\\
        & \textsc{sent-25} & 82.8 & 73.8 & 77.3 & 78.9 & 59.4 & 62.8 & 95.6 & 60.0 & 64.4 & 85.8 & 64.4 & 68.2\\\hdashline
        \multirow{3}{4em}{\IndE} %
        & \textsc{sent-75} & 80.2 & 75.9 & 77.8 & 82.8 & 78.2 & 80.3 & 46.7 & 50.0 & 48.3 & 70.0 & 68.9 & 69.1\\
        & \textsc{sent-50} & 82.6 & 79.4 & 80.9 & 85.9 & 80.9 & 83.2 & 45.4 & 50.0 & 47.5 & 71.3 & 70.1 & 70.5\\
        & \textsc{sent-25} & 85.9 & 76.5 & 80.2 & 77.4 & 80.1 & 78.7 & 70.6 & 85.4 & 74.5 & 77.9 & 79.9 & 77.5\\
        \bottomrule
        \end{tabular}
    \end{adjustbox}
    \caption{Performance comparison of models trained on samples created based on sentiment distribution. $\mu$ is the average
performance computed across all models.}
    \label{tab:pol_detailed}
\end{table*}

\section{Discussion}
Our findings from building a sentiment classification benchmark for linguistic varieties of English yield several actionable insights for the design of future benchmarks aimed at improving model robustness and generalisability. Here, we provide specific recommendations based on our evaluation of label semantics, text length variations, sentiment-bearing word proportions, and linguistic diversity.

\subsection{Impact of Label Semantics}
The significant performance drop observed when switching from \simple to \hard label semantics highlights the impact of label proximity on model performance. The \hard labels introduce ambiguity by narrowing the sentiment distinctions between reviews, making classification more challenging. While well-separated labels are beneficial for classification, we recommend incorporating more challenging closely grouped labels like \hard when creating a benchmark for sequence classification tasks.

\subsection{Incorporating Texts with Varying Lengths}
The analysis of length-based sampling revealed that models generally perform better on longer texts, likely due to the richer contextual information available in these samples. However, shorter reviews remain an important component of real-world data distributions and should not be overlooked. We recommend that future benchmarks explicitly incorporate both short and long text samples to create a balanced and challenging evaluation set. This approach ensures that models are assessed under different contextual conditions and varying levels of information.

\subsection{Managing Sentiment Proportion in Texts}
The performance degradation observed with sentiment distribution-based sampling, particularly for samples with a higher proportion of sentiment-bearing words, highlights the complexity introduced by this sampling strategy. In our methodology, sentiment distribution was calculated as the proportion of sentiment-bearing words (regardless of polarity) relative to the total number of words in a review. Consequently, a higher density of sentiment-bearing words does not necessarily correspond to a specific sentiment polarity.  Therefore, we recommend incorporating text with higher proportions of sentiment-bearing words, particularly when designing benchmarks for sentiment classification. This introduces ambiguity, making the task more challenging for models, as they cannot rely on an explicit sentiment signal.

\subsection{Language Variations and Generalisability}
The observed performance discrepancies across English varieties, with inner-circle varieties like en-AU and en-UK consistently yielding higher scores, highlight a bias toward high-resource language types. Addressing this bias is crucial for model robustness and generalisability. By incorporating text that features linguistic variations, cultural nuances, and underexplored dialects, benchmarks can more accurately reflect global linguistic diversity. This emphasis on variety forms the core motivation of our larger project to create a comprehensive benchmark that challenges models in cross-cultural and linguistically varied contexts.



\section{Related Work}
Traditional sentiment classification benchmarks like the Stanford Sentiment Treebank (SST-2)~\cite{socher-etal-2013-recursive} and the IMDB reviews dataset~\cite{maas-etal-2011-learning} focus on standard language forms and do not account for dialectal variations, limiting their applicability in multi-dialectal settings. Modern benchmarks such as GLUE~\cite{DBLP:journals/corr/abs-1804-07461} and SuperGLUE~\cite{DBLP:journals/corr/abs-1905-00537} aim for broader language understanding but still fall short in evaluating dialectal diversity. This paper aims to build upon the significant gap in the performance and evaluation practices of modern language models~\cite{joshi2024naturallanguageprocessingdialects}.
The creation of benchmarks that reflect linguistic diversity is crucial for ensuring that language technologies are equitable and do not perpetuate biases against specific linguistic subgroups~\cite{blodgett-etal-2020-language}. 

\citet{faisal2024dialectbench} propose DIALECT-BENCH, which contains NLP benchmarks for dialectal datasets but none for sentiment classification for dialects of English. To collect our dataset, we use a location-based selection of Google Place reviews. This is similar to past work in sentiment classification of dialectal Arabic~\cite{jurgens2017incorporating,muhammad2023afrisentitwittersentimentanalysis, boujou2021openaccessnlpdataset}. Similarly, ~\citet{coats-2022-corpus} use YouTube videos posted by Australian government departments as a heuristic for Australian English. To the best of our knowledge, ours is the first work to use Google Place reviews as the source of the dataset and the first to compare sentiment classification for Australian, Indian, and British reviews. We use the star rating provided by a review poster as the sentiment label. This is similar to past work which compares label proximity to the sentiment of a review~\cite{pang-lee-2005-seeing, maks-vossen-2013-sentiment}. Our sampling strategies that use length and sentiment distribution are similar to ~\citet{tang2015sentiment} and ~\citet{wang2013sample}.



\section{Conclusion \& Future Work}
This paper provides actionable insights for designing robust and challenging benchmarks for sentiment classification that account for linguistic diversity across English varieties. Our experiments reveal that the definition of label semantics, particularly the distinction between well-separated (\simple) and closely grouped (\hard) labels impacts model performance significantly. We also find that including a mix of short and long reviews enhances benchmark comprehensiveness, as models generally perform better with longer texts due to richer contextual information. However, incorporating higher proportions of sentiment-bearing words introduces ambiguity, making the task more challenging and revealing model limitations in understanding nuanced expressions. The discrepancies in model performances across language varieties underscore the need for benchmarks to include text beyond just inner-circle types, to improve model generalisability across different cultural and linguistic contexts. This work is part of our ongoing project to create a comprehensive benchmark for sequence classification tasks, including sentiment and sarcasm detection, with plans to expand data sources and add manual annotations to address the limitations of self-supervised labels and enhance the reliability of the dataset.

\section*{Limitations}
Not all reviews posted in Australia will be by speakers who self-identify as or use what is considered Australian English. This holds for India and the UK as well. However, because we report aggregate statistics overall reviews in the test sets that are posted within a locale, the numbers are notionally indicative of the performance of the models for these national varieties. We experiment with BERT-based models which may be considered a good baseline but not state-of-the-art. The benchmark described in the paper is a work in progress and will be released with additional annotations.

\section*{Ethical Considerations}
We treat the reviews are separate texts. We do not attempt to de-identify any users or collect aggregate information about individual users. The project received ethics approval from the human research ethics committee at ANONYMISED organisation.
\section*{Acknowledgment}
Will be added if the paper is accepted.

\bibliography{anthology}
\appendix
\onecolumn
\section{Annotation Exercise}\label{sec:annot}
This appendix details the annotation exercise conducted to assess the reliability and consistency of manual labels assigned to user-generated reviews, with a focus on distinguishing between various English varieties. Two annotators, ant-IN (a native speaker of Indian English) and ant-UK (a native speaker of British English), independently labelled the reviews as en-AU, en-IN, en-UK, or \textit{Cannot say}. The analysis highlights areas of both agreement and significant disagreement, illustrating the complexities of accurately labelling closely related English varieties and the challenges posed by linguistic nuances.
\begin{figure}[h!]
    \begin{adjustbox}{width=0.4\linewidth,center}
        \includegraphics[width=1\linewidth]{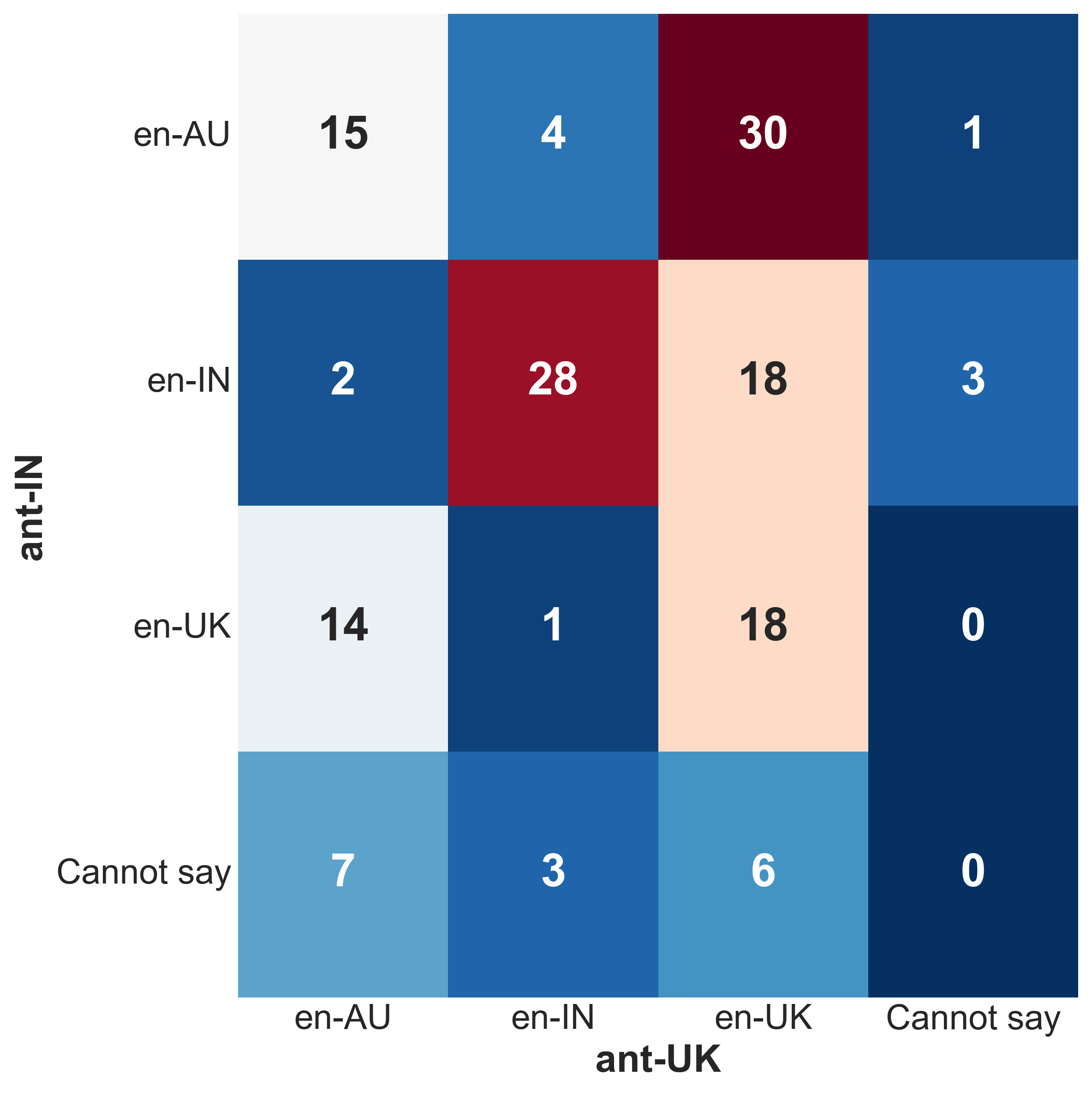}
    \end{adjustbox}
    \caption{Confusion Matrix of label annotations by ant-IN and ant-UK.}
    \label{fig:annot-mat}
\end{figure}
\subsection{Agreement Analysis}
Out of 150 reviews, the annotators agreed on 62 instances, with the highest agreement for the en-AU and en-IN labels. The Cohen’s Kappa score between ant-IN and ant-UK was 0.191, indicating slight agreement. ant-IN’s labels aligned more closely with the true reference labels, achieving a Kappa score of 0.506 (moderate agreement), while ant-UK’s labels had a Kappa score of 0.329 (fair agreement). ant-IN tended to favour en-IN (34.67\%) and en-AU (32.67\%), whereas ant-UK preferred en-UK (48.00\%) and en-AU (25.33\%). Both annotators used the \textit{Cannot say} label sparingly, reflecting a high level of confidence in assigning specific varieties, despite the overall low agreement. These variations underscore the subjective nature of labelling English varieties, influenced by each annotator’s background.

\subsection{Disagreement Analysis}
The analysis identified 88 instances of disagreement between ant-IN and ant-UK. As shown in Figure~\ref{fig:annot-mat}, 44 out of these 88 cases (50\%) stemmed from confusion between the en-AU and en-UK labels. ant-IN often labelled reviews as en-AU when ant-UK chose en-UK (30 instances) and vice versa (14 instances). This highlights the difficulty of distinguishing en-AU from en-UK, likely due to overlapping linguistic features.

The difficulty in distinguishing locales, even by human annotators, suggests that place-based reviews may not accurately capture language varieties as effectively as other data sources. This realisation has motivated our exploration of alternative data sources for our ongoing project of developing a comprehensive benchmark for sentiment and sarcasm classification across English varieties.

\end{document}